\documentclass[sigconf, nonacm]{acmart}
\AtBeginDocument{%
  }

\setcopyright{acmlicensed}
\copyrightyear{2018}
\acmYear{2018}
\acmDOI{XXXXXXX.XXXXXXX}
\acmConference[Conference acronym 'XX]{Make sure to enter the correct
  conference title from your rights confirmation email}{June 03--05,
  2018}{Woodstock, NY}
\acmISBN{978-1-4503-XXXX-X/2018/06}

\usepackage{xcolor}
\definecolor{agrcolor}{RGB}{142, 210, 107}
\definecolor{clicolor}{RGB}{216, 229, 244}
\definecolor{urbcolor}{RGB}{230, 195, 231}
\definecolor{forcolor}{RGB}{231, 93, 64}
\definecolor{satcolor}{RGB}{241, 184, 53}
\definecolor{datcolor}{RGB}{255, 251, 205}
\definecolor{dbcolor}{RGB}{242, 219, 207}
\definecolor{mapcolor}{RGB}{214, 239, 202}

\usepackage{tcolorbox}
\usepackage{multirow}

\begin{document}

\title[Geo-OLM: Sustainable Earth Observation with Open Language Models]{Geo-OLM: Enabling Sustainable Earth Observation Studies with Cost-Efficient Open Language Models \& State-Driven Workflows}


\author{Dimitrios Stamoulis}
\email{dstamoulis@utexas.com}
\orcid{0000-0003-1682-9350}
\affiliation{%
  \institution{Chandra Family Department of\\Electrical and Computer Engineering}
  \institution{The University of Texas at Austin}
  \city{Austin}
  \state{TX}
  \country{USA}}

\author{Diana Marculescu}
\email{dianam@utexas.edu}
\orcid{0000-0002-5734-4221}
\affiliation{%
  \institution{Chandra Family Department of\\Electrical and Computer Engineering}
  \institution{The University of Texas at Austin}
  \city{Austin}
  \state{TX}
  \country{USA}}


\begin{abstract}
Geospatial Copilots hold immense potential for automating Earth observation (EO) and climate monitoring workflows, yet their reliance on large-scale models such as GPT-4o introduces a \textit{\textbf{paradox}: tools intended for sustainability studies often incur unsustainable costs}. Using agentic AI frameworks in geospatial applications can amass thousands of dollars in API charges or requires expensive, power-intensive GPUs for deployment, creating barriers for researchers, policymakers, and NGOs. Unfortunately, when geospatial Copilots are deployed with open language models (OLMs), performance often degrades due to their dependence on GPT-optimized logic. In this paper, we present Geo-OLM, a tool-augmented geospatial agent that leverages the novel paradigm of state-driven LLM reasoning to decouple task progression from tool calling. By alleviating the workflow reasoning burden, our approach enables low-resource OLMs to complete geospatial tasks more effectively. When downsizing to small models below 7B parameters, Geo-OLM outperforms the strongest prior geospatial baselines by 32.8\% in successful query completion rates. Our method performs comparably to proprietary models achieving results within 10\% of GPT-4o, while reducing inference costs by two orders of magnitude from \$500–\$1000 to under \$10. We present an in-depth analysis with geospatial downstream benchmarks, providing key insights to help practitioners effectively deploy OLMs for EO applications.
\end{abstract}

\begin{teaserfigure}
  \includegraphics[width=\textwidth]{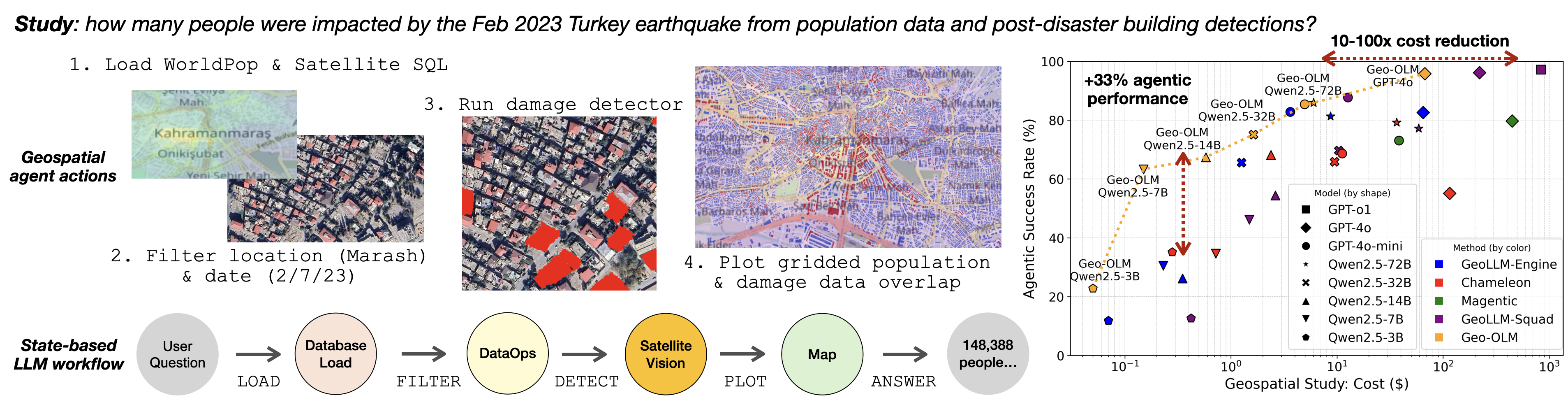}
  \caption{Earth observation studies, such as post-disaster societal impact assessments~\cite{robinson2023turkey}, involve complex workflows with multiple tools, data sources, and analytical steps. Automating such tasks with off-the-shelf Copilots -- particularly agents using proprietary models such as GPT-4o -- can rapidly escalate costs, making large-scale spatiotemporal assessments unsustainable for practitioners and researchers. We introduce Geo-OLM, a geospatial agentic framework based on state-driven prompting, where reasoning is structured into distinct stages. By decoupling task progression from tool calling, Geo-OLM enables competitive agentic performance with low-resource open language models while substantially reducing costs. \href{https://github.com/dstamoulis/geo-olms}{\textcolor{magenta}{[Project repo]}}} 
  \vspace{10pt}
  \Description{Overview of proposed methodology.}
  \label{fig:teaser}
\end{teaserfigure}



\maketitle


\section{Introduction}

Agentic AI has emerged as a powerful tool in geospatial sciences, where AI Copilots have the potential to transform Earth observation (EO) workflows~\cite{chen2024geoagent}. EO-based studies, such as climate and environmental monitoring, are inherently complex, requiring the processing of vast datasets, integration of multiple geospatial APIs, visualization through mapping systems~\cite{globalfishingwatch, nasa_earthdata_search}, and implementation of analytical techniques to identify long-term spatiotemporal trends~\cite{paolo2024satellite}. These tasks demand domain expertise in geosciences and are critical for sustainability research, making them prime candidates for automation through agentic AI~\cite{szwarcman2024prithvi}. 

Geospatial agents -- tool-augmented Large Language Models (LLMs) -- have demonstrated their ability to support climate studies and agriculture studies~\cite{li2024metafruit, yang2024multimodal}, urban planning~\cite{bhandari2024urban, yu2024harnessing}, and disaster response~\cite{goecks2023disasterresponsegpt} by automating data processing~\cite{chen2024geoagent}, visualization~\cite{zhang2024context}, and analysis. However, a fundamental \textbf{paradox} arises: while these tools are positioned as key enablers of sustainability research, their reliance on large-scale proprietary models incurs significant computational and financial costs, ultimately compromising their own sustainability. This challenge arises from prevailing agentic AI methodologies, which often prioritize maximizing Copilot performance over computational efficiency and cost-effectiveness~\cite{fourney2024magentic}.

To manage the complexity of geospatial applications, advanced LLM reasoning frameworks with multi-agent orchestration~\cite{fourney2024magentic, lu2024chameleon} rely on large models like GPT-4o, whose usage is prohibitively expensive. Conducting real-world geoscientific studies over satellite imagery spanning multiple terabytes -- requiring iterative querying, multi-step prompting, and extended user interactions -- can quickly accumulate costs in the thousands of dollars. Deploying open-source LLMs also presents significant challenges, as applying these agentic schemes to open language models (OLMs) results in severe performance degradation~\cite{lee2025geosquad}. Moreover, scaling with large OLMs (e.g., LLaMA-3.3 70B) demands A100-class GPUs, which themselves are costly and have a high carbon footprint. This raises critical questions about accessibility, particularly for researchers, NGOs, and policymakers -- stakeholders who stand to benefit most from geospatial AI but might be least equipped to bear these costs. 

To address the need for sustainable agentic EO applications, our goal is to enable geospatial agents that operate effectively with small and mid-sized OLMs, typically models with 14B parameters or fewer~\cite{abdin2024phi}. However, at this smaller scale, OLMs struggle with the complex reasoning required for geospatial tasks. To this end, we draw the attention to 
an aspect often underleveraged in agentic AI design: unlike generalist Copilots designed for broader tasks like web browsing or document summarization~\cite{zhou2024webarena}, geospatial workflows follow a structured, sequential progression~\cite{singh2024evalrs}. Consider a geoscientist: \textit{they must first load and preprocess data before performing filtering, and similarly, data analysis must be completed before generating map visualizations}. This inherent structure presents us with a key opportunity: by explicitly modeling this progression, we can design more efficient agentic prompting that align with geospatial task dependencies and is better suited for smaller OLMs.

In this work, we present Geo-OLM, geospatial agents that leverage the recently proposed state-driven LLM prompting paradigm~\cite{wu2024stateflow} to alleviate the reasoning burden in low-resource OLMs. Our approach models geospatial workflows as discrete states and transitions, decoupling process grounding and task execution from tool calling and sub-task solving. Each state is described by its expected functionalities and supplemented with few-shot examples to facilitate tool selection, so transitions can be guided by lightweight LLM-based logic~\cite{wu2024stateflow}. This mitigates the computational overhead of multi-round agentic orchestration logic, enabling smaller OLMs to achieve higher task completion rates at significantly lower cost.

We evaluate our approach on state-of-the-art geospatial agentic benchmarks, GeoLLM-Engine~\cite{singh2024geoengine} and GeoLLM-Squad~\cite{lee2025geosquad}, which encompass single- and multi-agent EO workflows, respectively. We conduct a comprehensive evaluation across various prompting schemes and LLM APIs, including proprietary GPT models and open LLM families (e.g., LLaMA~\cite{dubey2024llama}, Phi~\cite{abdin2024phi}, Qwen~\cite{yang2024qwen2}). Our analysis spans multiple API endpoints such as OpenAI and  Ollama~\cite{ollama}. To our knowledge, this is the first work to leverage state-driven prompting within OLMs for geospatial tasks. 

Our results show that, when tested on proprietary models like GPT-4o, Geo-OLM achieves performance within 1\% of state-of-the-art methods. When applied to smaller OLMs, our approach maintains robust performance at a fraction of the cost, reducing expenses from over \$1,000 to less than \$10 on EO benchmarks -- a 100x decrease. More importantly, at that downsized scale with models of 14B parameters or less (e.g., Qwen-2.5-7B~\cite{yang2024qwen2}), existing geospatial solutions experience a performance collapse by up to 60\% in completion rates, while Geo-OLM remains within 10-20\% of the large baselines. Overall, state-driven reasoning offers substantial cost benefits, achieving near-GPT performance. We demonstrate the generalizability of Geo-OLM beyond benchmark scenarios through a case study on a real-world environmental analysis of the 2023 Turkey earthquake~\cite{robinson2023turkey}.

Our main contributions are as follows: (1) We introduce Geo-OLM, a geospatial prompting scheme that models EO workflows as state machines, enabling enhanced control and efficiency for EO tasks using smaller OLMs. (2) We conduct an extensive study across major LLM serving runtimes (e.g., Ollama, OpenAI) and model families (Phi, Qwen, LLaMA), offering practitioner guidelines for optimizing geospatial agentic prompting. (3) We evaluate our approach on both single- and multi-agent tasks and report on both monetary/cloud costs and task success rates, providing a holistic view of agentic performance. We will release our code and prompts upon acceptance to facilitate further research and adoption. We hope that our method will serve as a valuable resource for geoscientists and practitioners seeking cost-efficient and sustainable agentic practices for EOw studies.

\begin{figure*}[t]
  \centering
  \includegraphics[width=\linewidth]{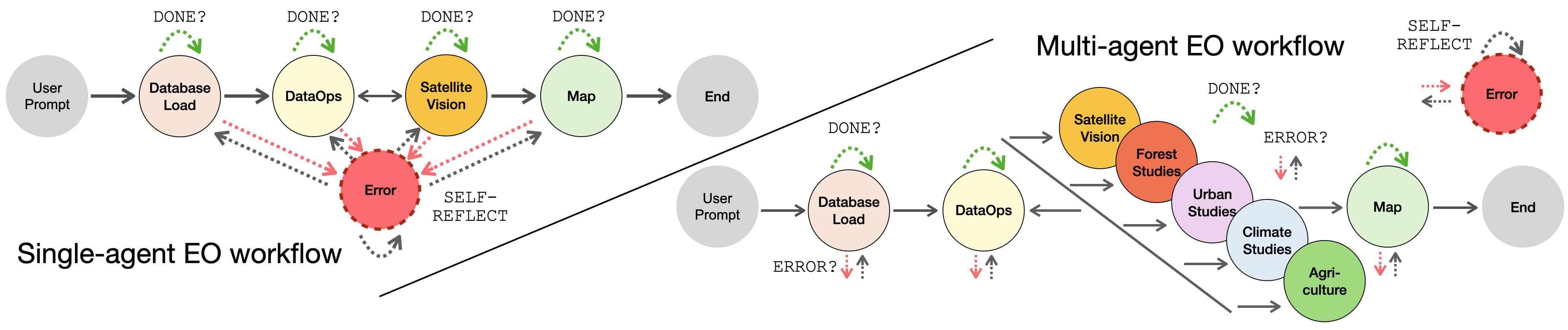}
  \caption{State-driven LLM reasoning with Geo-OLM. Earth observation (EO) workflows exhibit inherent progression logic, which we encapsulate using the StateFlow paradigm~\cite{wu2024stateflow}. We illustrate this first in a single-agent scenario for satellite detection in remote sensing (RS) tasks (left) and extend it flexibly to a multi-agent setup incorporating diverse EO applications (right). By structuring reasoning into explicit states, Geo-OLM reduces the cognitive burden on low-resource OLMs, allowing them to better reflect on task progression, correctly handle errors, and prevent premature function termination.}
  \label{fig:eo_workflows}
\end{figure*}

\section{Related work}

\textbf{Geospatial Foundation Models (GFMs)}. There is rapid progress in applying generative AI advances to EO tasks~\cite{szwarcman2024prithvi}, with recent works demonstrating their potential across various domains, including satellite vision~\cite{dias2024oreole, mall2024remote, hu2023rsgpt, silva2024large, Liu2024remoteclip, Zhang2024EarthGPT, jian2023stable}, urban analysis~\cite{bhandari2024urban, yu2024harnessing}, map-based applications~\cite{zhang2024context}, agriculture\cite{li2024metafruit, yang2024multimodal, microsoft2023agriculture, bountos2023fomo, lacoste2024geo}, and forestry~\cite{zhu2024foundations, xiong2024neural}. However, these approaches typically focus on a narrowly defined scope, where a fine-tuned LLM is trained for a specific task within a predetermined workflow~\cite{hu2023rsgpt, Zhang2024EarthGPT}. While effective, this approach is akin to deploying task-specific computer vision models, requiring extensive labeled data and computational resources for fine-tuning. In contrast, there is growing interest in more generalizable, fine-tuning-free solutions with LLMs.

\textbf{Geospatial agents}: A promising approach for geospatial AI is function-calling via LLMs, where an agent (tool-augmented LLM) interprets natural language user queries and selects the appropriate tools to complete a task. Recent geospatial copilots~\cite{singh2024geoengine, chen2024geoagent} extend this capability to multi-scope EO tasks, integrating various APIs to support complex, long-horizon workflows. However, these systems centralize the prompting, decision-making, reasoning, and tool execution logic, making them reliant on large-scale proprietary models to function effectively~\cite{lee2025geosquad}. This issue is even more pronounced in generalist frameworks, where the focus is on maximizing performance at the expense of efficiency. For instance, composition-based~\cite{lu2024chameleon} and ledger-based scheduling~\cite{fourney2024magentic} paradigms require extensive message passing and iterative reasoning across multiple LLM calls to converge on a task. These computationally expensive techniques often make state-of-the-art methods cost-prohibitive for deployment in sustainability and climate studies.

\textbf{Open Language Models}. In the last few weeks alone, we have witnessed rapid advances in smaller open and open-weights models that train faster, are optimized for low latency~\cite{mistral2025small}, and increasingly match larger proprietary models in coding tasks and tool-calling workflows~\cite{yang2024qwen2}. Breakthroughs such as the Phi model family~\cite{abdin2024phi} and DeepSeek models~\cite{guo2025deepseek} demonstrate impressive potential toward closing the performance-cost gap. Similarly, the open-source ecosystem for OLM deployment is maturing rapidly, with community-driven initiatives like vLLM~\cite{vllm2025, kwon2023efficient} and more commercially focused platforms like LangChain~\cite{langchain2024docs} and Ollama~\cite{ollama}.

However, these platforms remain primarily focused on generalist tasks, making them not directly applicable to workflows requiring domain expertise. Additionally, the lack of a widely established testbed for geospatial agentic tasks makes it difficult to compare model families or track improvements between older and newer models~\cite{zhang2024good}. Similar discrepancies in evaluation methodologies have been noted in other domains, such as biomedical agentic applications~\cite{lobentanzer2025biomedical}. In this work, we consider these challenges by conducting an in-depth evaluation across a wide range of the latest OLM variants from all major families.

\section{Methodology}

\subsection{Background: State-driven LLM reasoning}

In state-based LLM reasoning, the task-solving process is conceptualized as a finite state machine (FSM)~\cite{wagner2006modeling}, albeit a simplified version that depends on context history rather than an explicit input tape~\cite{wu2024stateflow}. In LLMs, this context is inherently provided by the chat history of agent-user interactions (i.e., the \texttt{messages = [..]} variable in LLM APIs) which accumulates as the execution progresses. By adapting an FSM-like notation, a workflow can be formulated as a tuple $\langle S, \delta, \Gamma \rangle$, where: (1) $S$ represents the set of states, each corresponding to a distinct stage in the workflow, typically linked to executing a specific sub-task (e.g., loading data, applying filters, generating visualizations). (2) $\delta$ defines state transitions, determining the next state based on the current execution context and message history. (3) $\Gamma$ denotes the sequence of messages exchanged during execution, including prompts, tool definitions, invocations, and responses.

More concretely, over $n$ LLM API calls, $\Gamma$ consists of the initial task prompt $P$, agent instructions $I = \{I_1, I_2, \dots, I_n\}$, tool definitions $T = \{T_1, T_2, \dots, T_n\}$ specifying the tools available at each stage, LLM responses $R = \{R_1, R_2, \dots, R_n\}$, and tool return messages $O = \{O_1, O_2, \dots, O_n\}$. In chat-based LLM APIs, these components have dedicated roles: \texttt{system} (task prompt $P$), \texttt{user} (instructions $I$), \texttt{tools} argument (tool definitions $T$), \texttt{assistant} (LLM responses $R$), and the recently introduced \texttt{tool/ipython} (tool outputs $O$). 

State transitions $\delta$ operate as a mapping function $(\delta : S \times \Gamma \rightarrow S)$, determining the next state based on the current state and accumulated message history. In practice, the prompt engineer defines the states, as well as the available tools per execution states, the expected actions, and tool invocations directly into LLM prompts. Execution logic is implemented as a combination of simple heuristics (e.g., programmatically verifying SQL query execution before proceeding) and LLM-based reasoning.

\subsection{State-driven EO workflows}

We begin by identifying a representative workflow for EO tasks, specifically defining the states ($S$) and state transitions ($\delta$). We use Remote sensing (RS) tasks from the GeoLLM-Engine benchmark to illustrate the process (we generalize to other EO tasks in the following Subsection). GeoLLM-Engine consists of multi-round tasks, such as: ``\textit{From the xView1 satellite imagery, run vessel detection and plot all ships from the Port of Istanbul in May 2020.}'' 

\textbf{Overall flow}. As shown in Figure~\ref{fig:eo_workflows} (left), the process in such an EO workflow begins with loading the specified satellite product from the SQL database. Next, the data is filtered based on user criteria (e.g., temporal and spatial constraints such as date ranges and geographic coordinates). The agent then executes the detection model on the satellite imagery. Depending on the task requirements, additional refining steps may follow -- for instance, filtering detections based on specific object categories. Finally, the system renders the detections onto a map and returns the results to the user.

\textbf{States}. To structure execution within the LLM, we define the state sequence for single-task EO queries as \texttt{Init} $\rightarrow$ \texttt{Load} $\rightarrow$  \texttt{Filter} $\rightarrow$ \texttt{Detect} $\rightarrow$ \texttt{Map} $\rightarrow$ \texttt{End}. In each state, the model is instructed to perform a set of recommended actions corresponding to its designated stage in the workflow.

\textbf{LLM prompting}. The LLM is first provided with an explicit definition of the workflow, outlining the structured progression of states. Then, at each interaction step, the model is instructed to infer and explicitly output its current execution stage ($S$) based on the accumulated messages history. This is implemented by requiring the LLM to append \texttt{CURRENT\_STAGE = ...} to its response, ensuring that the model continuously self-identifies its position in the workflow.

\textbf{Per-state tools}. At each state, we associate the set of tools that are relevant to completing the corresponding step. Tool definitions are appended to the model input (via the tools API argument) and the LLM responds with the selected function calls to execute. While this follows the standard LLM function-calling mechanism, our state-driven approach introduces a key efficiency gain: since the LLM explicitly tracks its execution state, we can dynamically suggest tools relevant for each stage. For example, if the \texttt{CURRENT\_STAGE} is identified as \texttt{Load}, the agent is programmatically provided with few-shot examples related to database query functions. This selective prompting significantly reduces decision complexity, as also motivated in more elaborate dynamic tooling schemes~\cite{liu2024tool, paramanayakam2024less, srinivasan2023nexusraven}, as it is beneficial for smaller OLMs which might struggle when presented with a large set of tool choices.

\textbf{Transitions}. As in standard LLM execution, transitions between workflow stages are LLM-driven, where the model infers the appropriate next step based on the accumulated message history and updates \texttt{CURRENT\_STAGE} accordingly. However, conceptualizing the workflow as a state machine introduces two key advantages that can further reduce the reasoning burden on smaller OLMs: structured error handling and task completion validation.

For \textbf{\textit{error}} handling, we introduce a dedicated \texttt{Error} state to manage failures. Instead of treating execution errors as unstructured chat history (as in conventional function-calling pipelines), state-driven reasoning enables programmatic transitions to the Error state when an exception is detected. Upon entering this state, the system issues a standalone LLM call with explicit \texttt{SELF-REFLECT} instructions, prompting the model to reconsider its previous actions and identify potential errors. To assist in debugging, the LLM is provided with details of the last function call, execution results, and the corresponding tool definitions, allowing it to \textit{self-correct} minor mistakes such as typos or incorrect parameter usage.

For task \textbf{\textit{completion}}, we introduce a termination safeguard. This is done by instructing the LLM to append the keyword \texttt{TERMINATE} when it determines that the task is complete. This signal is programmatically parsed, and upon detection, a follow-up validation step is issued: the model is explicitly asked to reason and confirm whether execution has indeed concluded by summarizing its current output. This mechanism helps prevent premature exits or infinite loops due to incorrect state reasoning or erroneous function calling.

\subsection{Expanding to other EO workflows}

State-based reasoning extends naturally to multi-agent flows~\cite{wu2024stateflow}, enabling broader geospatial applications beyond the single-task RS examples discussed above. We expand the state sequence definitions to enumerate the different possible agentic functionalities in the LLM prompt -- as different paths shown in Figure~\ref{fig:eo_workflows} (right). This is implemented by requiring the LLM to interpret the original user task and append \texttt{USER\_INTENT = ...} to its response. The list of possible intents is flexibly adapted to the workflow of interest. In our study, we consider GeoLLM-Squad~\cite{lee2025geosquad}, a multi-scope geospatial designed for diverse satellite-based analytics for urban monitoring, agriculture, forest management, and climate analysis, and satellite detections (see next Section for benchmark details). To this end, we ask the LLM to reason and select from \texttt{Vision (Detect)}, \texttt{Forest}, \texttt{Urban}, \texttt{Climate}, and \texttt{Agriculture}.

\section{Geo-OLM framework}

Our Geo-OLM implementation consists of two main components: the agentic frontend, which handles all interactions between the user and the agent, and the LLM API serving backend, responsible for executing model queries. 

\textbf{Agentic frontend}. We build our frontend by customizing agentic routines from AutoGen~\cite{wu2023autogen}. As is common in agentic AI evaluations, we adopt a command-line sandbox implementation to ensure controlled execution, particularly when interacting with external tools such as web APIs or SQL databases. In this setting, the system records agent actions and executes the corresponding API calls in isolation, rather than requiring a full UI-driven workflow~\cite{dibia2024autogen}. 

This design significantly simplifies implementation, allowing us to focus on developing the API toolset necessary to solve GeoLLM benchmarks without UI integration overhead, while maintaining full execution fidelity~\cite{koh2024visualwebarena}. For function calling, we compile the list of functionalities and respective API tools required to solve the considered benchmarks~\cite{lee2025geosquad, singh2024geoengine}. We reimplement the logic (the specific tools, e.g., loading xview1 data) and we confirm correctness by closely matching metrics reported in the baselines used in our comparisons.

\textbf{LLM serving backend}. A key component of our implementation is the LLM runtime engine, which enables local execution of open LMs -- an aspect often overlooked in existing geospatial agents that rely solely on GPT-based APIs. We integrate the \texttt{Ollama}~\cite{ollama} LLM serving framework to support both text-based and chat-based model inference. For models that do not natively support tool execution within these APIs, we follow the official schema recommendations, specifying regex-based parsing to extract function calls based on tool role definitions. Otherwise, we leverage the built-in API schemas directly.

\textbf{Benchmarks}. To evaluate Geo-OLM, we conduct experiments on two benchmarks, namely GeoLLM-Engine~\cite{singh2024geoengine} and GeoLLM-Squad~\cite{lee2025geosquad}, which cover realistic EO tasks and different user workflows over millions of satellite images and large-scale remote sensing products: (1) GeoLLM-Engine consists of single-tasks, multi-step workflows for object detection, land cover classification, and visual question answering (VQA) with satellite imagery. The benchmark integrates open-source satellite imagery datasets with globally distributed spatiotemporal metadata, requiring agents to reason over specific locations and time periods. (2) GeoLLM-Squad covers a wider scope with multi-round EO workflows requiring interactions across different data modalities based on satellite products for urban analysis, agriculture and forest monitoring, and climate-related retrievals. Both benchmarks involve open-source datasets such as built-up area and population density mapping from Sentinel-2 imagery~\cite{hansen2013forestchange}, and vegetation indices or surface temperature data from NASA MODIS Terra products~\cite{nasa_earthdata_search, vermote2015reflectance, wan2021lst}. We standardize all products, including HDF and GeoTIFF formats, as \texttt{DataFrame} variables, facilitating seamless integration into our framework. 

\textbf{Case study}. Unlike existing works that focus exclusively on curated benchmarks, we extend our evaluation to real-world scenarios. To evaluate how Geo-OLMs generalize beyond predefined datasets, we replicate the analytical workflow of the February 2023 Turkey earthquake study~\cite{robinson2023turkey}. Using the open-source satellite-based building damage assessments and population estimates, we prompt Geo-OLMs to follow the original work's methodology via state-driven reasoning, assessing their agentic ability to reach the originally reported findings within reasonable fidelity.

\begin{table*}
  \caption{GPT models: Agentic correctness and success rates with cost and downstream EO task performance~\cite{lee2025geosquad}.}
  \label{tab:gpt_results}
  \centering
  \resizebox{\textwidth}{!}{ 
  \begin{tabular}{l c c c c r r r r r r r}
    \toprule
    \multirow{2}{*}{\textbf{Model}} & \textbf{Geospatial}  & \textbf{Chat} & \textbf{Correct.} & \textbf{Success} & \textbf{Tokens} & \textbf{Total} & \textbf{Agro} & \textbf{Climate} & \textbf{Urban} & \textbf{Forest} & \textbf{Vision} \\
     & \textbf{Agent} &  \textbf{API} & \textbf{Rate}\% & \textbf{Rate}\% & \textbf{Avg (k)} & \textbf{Cost} & $\epsilon$\% & $\epsilon$\% & $\epsilon$ \% & $\epsilon$\% & \textbf{R}\% \\
    \midrule
    GPT-o1 & GeoLLM-Squad  & \checkmark & 90.97 & 97.2 & 77.36 & \$833.91 & 0.14 & 0.58 & 1.39 & 0.56 & 98.71 \\
    \midrule
    \multirow{6}{*}{GPT-4o} & GeoLLM-QA & \checkmark & 52.49 & 62.3 & 22.78 & \$68.23 & 5.34 & 3.97 & 6.33 & 6.73 & 72.80 \\
     & GeoLLM-Engine  & \checkmark & 74.11 & 82.6 & 22.25 & \$64.72 & 5.05 & 2.26 & 8.31 & 7.85 & 87.71 \\
     & Chameleon  & \checkmark & 21.87 & 55.1 & 36.19 & \$115.08 & 12.51 & 9.14 & 9.62 & 10.77 & 50.42 \\
     & Magentic  & \checkmark & 25.40 & 79.7 & 117.65 & \$447.55 & 83.89 & 80.63 & 80.86 & 83.13 & 97.25 \\
     & GeoLLM-Squad  & \checkmark & 85.32 & 96.2 & 77.49 & \$219.67 & 3.70 & 2.02 & 1.99 & 3.33 & 97.59 \\\cmidrule(lr){2-12}
     & Geo-OLMs & \checkmark & 86.03 & 95.1 & 30.44 & \$77.02 & 3.44 & 1.17 & 4.40 & 2.51 & 96.26 \\
    \midrule
    \multirow{6}{*}{GPT-4o-mini} & GeoLLM-QA  & \checkmark & 37.41 & 53.1 & 21.55 & \$3.99 & 13.05 & 10.29 & 14.61 & 12.82 & 69.16 \\
     & GeoLLM-Engine  & \checkmark & 53.56 & 82.8 & 20.09 & \$3.63 & 11.45 & 8.69 & 8.88 & 8.59 & 60.33 \\
     & Chameleon & \checkmark & 34.63 & 68.7 & 60.10 & \$11.17 & 12.25 & 8.05 & 8.85 & 8.84 & 67.42 \\
     & Magentic  & \checkmark & 18.85 & 73.1 & 187.45 & \$38.33 & 13.44 & 11.45 & 11.54 & 12.29 & 75.51 \\
     & GeoLLM-Squad  & \checkmark & 59.46 & 87.7 & 72.98 & \$12.63 & 16.42 & 9.70 & 10.22 & 10.19 & 70.58 \\\cmidrule(lr){2-12}
     & Geo-OLMs  & \checkmark & 58.75 & 85.4 & 28.49 & \$4.96 & 15.66 & 10.45 & 9.75 & 8.63 & 75.49 \\
    \bottomrule
  \end{tabular}
  }
\end{table*}

\section{Experimental Setup}

\textbf{Model families}. To our knowledge, this is the first geospatial agentic study to systematically profile both proprietary and open LLMs to this extent, as we consider numerous major tool-supporting APIs available in \texttt{Ollama} as of January 2025. For proprietary models, we use OpenAI’s latest GPT-4o, GPT-4o-mini, and GPT-o1, while for open models we report results for the LLaMA-3.3 and 3.2 series (70B, 3B, 1B)~\cite{dubey2024llama}, the Qwen-2.5 family (72B down to 0.5B)~\cite{yang2024qwen2}, Mistral (Small 14B)~\cite{mistral2025small}, and the latest Phi-4 (14B)~\cite{abdin2024phi}. 

\textbf{LLM baselines}. We evaluate both generic and geospatial-specific LLM prompting methods. For comparison purposes, we reimplement the GeoLLM-Engine and GeoLLM-Squad solvers within our framework, which use ReAct~\cite{yao2022react} and Orchestrator~\cite{fourney2024magentic} prompting, respectively. Moreover, we compare with Chameleon~\cite{lu2024chameleon}, a compositional scheduling method, and Magentic, a generalist multi-agent framework~\cite{fourney2024magentic}. We confirm our implementations closely match reported values from prior work.

\textbf{Metrics}. For performance on EO tasks, we report metrics as in the baseline benchmarks, e.g., F1 scores for object detection tasks. To assess agentic execution, we adopt widely-used evaluation practices from WebArena~\cite{zhou2024webarena, koh2024visualwebarena, workarena2024}, computing \textit{success} and \textit{correctness} rates. Success rate measures the proportion of fully completed tasks, regardless of intermediate errors, while correctness rate quantifies the fraction of correct function calls with the correct parameters executed in the expected order.  

\textbf{Testbed}. GPT experiments use the OpenAI API, while all OLM evaluations are run on a dedicated GPU cluster equipped with 8 A series GPUs (40GB each). To ensure consistent cost measurements and reduce potential variability from network delays, we store all datasets locally on disk. We compute the average time and tokens per query (from initial user question to completion) over the entire benchmark (2k). All software dependencies use the latest available versions as of January 2025. For Ollama, we use the default Q4 quantized models. 

\textbf{Cost model}. We report the total cost of executing the full benchmark (2K queries). For GPT models, we log the exact breakdown of input, cached, and output tokens from the OpenAI API and compute costs using official pricing rates~\cite{openai_pricing2025}. For OLMs running locally, we estimate costs based on market-rate GPU pricing. Our setup is closest to the Azure NC48ads series~\cite{azure_nd_a100_2025}, with an estimated rate of \$10 per hour. For representative comparison with a real-world deployment (where the requests will be batched over a server), we amortize costs by the maximum number of models that can fit concurrently in the GPUs based on Ollama memory usage. We compute OLM costs by multiplying this amortized hourly rate by the profiled per-query inference time.

\begin{table*}
  \caption{Qwen-2.5 models: Agentic correctness and success rates with cost and downstream EO task performance~\cite{lee2025geosquad}.}
  \label{tab:olms_qwen_results}
  \centering
  \resizebox{\textwidth}{!}{ 
  \begin{tabular}{l l c c c r r r r r r r}
    \toprule
    \multirow{2}{*}{\textbf{Model}} & \textbf{Geospatial}  & \textbf{Chat} & \textbf{Correct.} & \textbf{Success} & \textbf{Tokens} & \textbf{Total} & \textbf{Agro} & \textbf{Climate} & \textbf{Urban} & \textbf{Forest} & \textbf{Vision} \\
     & \textbf{Agent} &  \textbf{API} & \textbf{Rate}\% & \textbf{Rate}\% & \textbf{Avg (k)} & \textbf{Cost} & $\epsilon$\% & $\epsilon$\% & $\epsilon$ \% & $\epsilon$\% & \textbf{R}\% \\
    \midrule
    \multirow{6}{*}{Qwen-2.5-72B} & GeoLLM-Engine (GE) &  & 45.06 & 81.3 & 26.43 & \$8.69 & 22.08 & 20.73 & 20.05 & 19.14 & 96.24 \\
     & GE+Our \texttt{ERR/TRM} &  \checkmark & 51.32 & 84.9 & 29.39 & \$5.79 & 13.06 & 11.65 & 8.80 & 8.93 & 85.69 \\
     & Chameleon &  \checkmark & 25.74 & 79.2 & 99.36 & \$36.48 & 16.13 & 12.89 & 11.52 & 10.45 & 46.39 \\
     & GeoLLM-Squad (GS) &  & 45.36 & 77.2 & 85.28 & \$58.83 & 32.44 & 29.28 & 29.93 & 28.78 & 96.68 \\
     & GS+Our \texttt{ERR/TRM} &  \checkmark & 56.39 & 85.4 & 82.34 & \$53.40 & 11.28 & 11.01 & 10.96 & 9.68 & 92.67 \\\cmidrule(lr){2-12}
     & Geo-OLMs &  \checkmark & 57.52 & 85.9 & 33.22 & \$6.01 & 11.84 & 11.27 & 10.65 & 9.71 & 94.53 \\
    \midrule
    \multirow{6}{*}{Qwen-2.5-32B}  & GeoLLM-Engine (GE) &  & 46.48 & 65.6 & 27.11 & \$1.26 & 13.77 & 9.84 & 10.36 & 10.70 & 91.24 \\
     & GE+Our \texttt{ERR/TRM} &  \checkmark & 48.24 & 73.6 & 25.34 & \$1.71 & 14.06 & 10.67 & 9.91 & 9.16 & 87.60 \\
     & Chameleon &  \checkmark & 23.77 & 65.9 & 99.13 & \$9.45 & 15.37 & 10.27 & 10.00 & 8.45 & 75.96 \\
     & GeoLLM-Squad (GS) &  & 52.40 & 69.7 & 76.91 & \$10.39 & 13.83 & 10.00 & 10.69 & 9.46 & 88.75 \\
     & GS+Our \texttt{ERR/TRM} &  \checkmark & 56.03 & 75.4 & 81.58 & \$13.64 & 13.78 & 10.15 & 9.95 & 8.74 & 88.28 \\\cmidrule(lr){2-12}
     & Geo-OLMs &  \checkmark & 55.91 & 75.1 & 29.71 & \$1.63 & 13.63 & 10.62 & 10.21 & 9.25 & 86.67 \\
    \midrule
    \multirow{6}{*}{Qwen-2.5-14B}  & GeoLLM-Engine (GE) &  & 31.93 & 26.2 & 29.68 & \$0.35 & 39.32 & 35.83 & 31.71 & 29.73 & 82.13 \\
     & GE+Our \texttt{ERR/TRM} &  \checkmark & 37.97 & 63.8 & 31.37 & \$0.46 & 17.92 & 13.33 & 11.84 & 11.29 & 80.26 \\
     & Chameleon &  \checkmark & 22.65 & 68.2 & 106.13 & \$2.38 & 16.21 & 12.90 & 11.38 & 11.12 & 86.86 \\
     & GeoLLM-Squad (GS) &  & 42.99 & 54.4 & 86.22 & \$2.63 & 39.91 & 33.88 & 29.93 & 28.89 & 90.41 \\
     & GS+Our \texttt{ERR/TRM} &  \checkmark & 44.09 & 67.4 & 89.43 & \$3.47 & 19.90 & 15.13 & 13.56 & 13.90 & 81.71 \\\cmidrule(lr){2-12}
     & Geo-OLMs &  \checkmark & 42.79 & 67.4 & 37.08 & \$0.58 & 19.43 & 15.05 & 13.51 & 14.16 & 87.86 \\
    \midrule
    \multirow{6}{*}{Qwen-2.5-7B}  & GeoLLM-Engine (GE) &  & 19.84 & 30.5 & 55.52 & \$0.23 & 50.44 & 45.62 & 43.39 & 41.66 & 72.16 \\
     & GE+Our \texttt{ERR/TRM} &  \checkmark & 26.75 & 56.2 & 51.02 & \$0.23 & 19.42 & 15.13 & 14.39 & 13.90 & 48.27 \\
     & Chameleon &  \checkmark & 14.64 & 34.6 & 99.92 & \$0.72 & 98.37 & 94.69 & 94.77 & 93.57 & 97.59 \\
     & GeoLLM-Squad (GS) &  & 33.96 & 46.2 & 149.54 & \$1.49 & 31.00 & 27.50 & 25.82 & 22.96 & 51.91 \\
     & GS+Our \texttt{ERR/TRM} &  \checkmark & 36.63 & 60.5 & 153.21 & \$2.74 & 15.86 & 13.00 & 12.64 & 11.22 & 66.36 \\\cmidrule(lr){2-12}
     & Geo-OLMs &  \checkmark & 40.90 & 63.3 & 40.77 & \$0.15 & 15.79 & 13.00 & 12.59 & 10.50 & 59.98 \\
    \midrule
     \multirow{6}{*}{Qwen-2.5-3B} & GeoLLM-Engine (GE) &  & 9.42 & 11.8 & 52.01 & \$0.07 & 103.08 & 97.25 & 89.65 & 87.08 & 89.55 \\
     & GE+Our \texttt{ERR/TRM} &  \checkmark & 19.26 & 24.9 & 45.84 & \$0.06 & 23.59 & 20.49 & 19.36 & 18.96 & 49.85 \\
     & Chameleon &  \checkmark  & 9.09 & 35.1 & 108.50 & \$0.28 & 17.43 & 13.82 & 12.05 & 12.21 & 52.58 \\
     & GeoLLM-Squad (GS) &  & 7.94 & 12.6 & 146.31 & \$0.42 & 92.05 & 87.29 & 82.95 & 82.28 & 92.92 \\
     & GS+Our \texttt{ERR/TRM} &  \checkmark & 25.63 & 19.5 & 149.85 & \$0.48 & 23.33 & 19.47 & 18.72 & 18.46 & 74.72 \\\cmidrule(lr){2-12}
     & Geo-OLMs &  \checkmark & 26.26 & 22.8 & 50.87 & \$0.05 & 28.97 & 25.62 & 24.10 & 23.91 & 80.89 \\
    \bottomrule
  \end{tabular}}
\end{table*}

\section{Results}

\textbf{GPT Models}. To assess the reasoning effectiveness of state-driven prompting, we first compare Geo-OLM against existing GPT-based geospatial methods with the latest GPT-4o and o1 variants (Table~\ref{tab:gpt_results}). Among the GPT-4o models, Geo-OLM achieves agentic rates within 1\% from the strongest prior work, GeoLLM-Squad, and within 2-4\% of the near-oracle GPT-o1, while reducing costs by an order of magnitude (\$77.02 vs. \$833.91). Similar trends hold for GPT-4o-mini, where Geo-OLM achieves an 85.4\% success rate, within 2.3\% of GeoLLM-Squad, but at a 2.5 $\times$ lower cost. In terms of EO metrics, Geo-OLM GPT-4o has lower error rates across most applications, while performance with GPT-4o-mini is stronger for satellite detections and within 2\% of GeoLLM-Engine $\epsilon$ rates. Last, confirming our implementation, the correctness rates for the baseline methods with GPT-4o-mini are within 2\% of previously reported values~\cite{lee2025geosquad}.

\textbf{OLMs - Qwen family}. Starting with the latest Qwen-2.5 variants (Table~\ref{tab:olms_qwen_results}), we evaluate agentic performance beyond proprietary LLMs. We observe that Geo-OLM achieves a better performance-cost trade-off across different models sizes (from 72B down to 3B). For the stronger variant (Qwen-2.5-72B), Geo-OLM achieves rates within 1–2\% of its GPT-4o-mini counterpart while running at comparable cost, demonstrating the feasibility of sustained geospatial analysis without reliance on opaque proprietary models.

\textbf{Prior geospatial agents with OLMs}. From Table~\ref{tab:olms_qwen_results}, we note how quickly agentic performance deteriorates as we move to low-resource OLMs. Both GeoLLM-Engine and GeoLLM-Squad experience a 30–40\% drop in completion rates at 7B, while Geo-OLM remains consistently higher by up to 33\%. Since we have previously verified our implementations by matching reported GPT-based metrics, we hypothesize that this degradation in existing geospatial agents stems from how prior work handles function calling with OLMs. Specifically, both methods report direct integration with text-based API logic rather than utilizing chat-based support (\texttt{tools} argument in HuggingFace/Ollama).

To investigate further, we conduct an in-depth failure analysis. We find that lower success rates in existing works primarily result from models exiting prematurely, while correctness errors often arise from incorrect tool argument formatting. For example, function calls with OLMs frequently fail due to minor syntax errors, such as passing \texttt{startdate} instead of \texttt{start\_date}. Additionally, the reliance on text-based logic introduces JSON parsing schemas on top of the model output, contributing further to typos and misformatting errors. These findings indicate that addressing these failures should improve performance.

\begin{table*}
  \caption{LLaMA, Mistral, and Phi models: Agentic correctness and success rates with cost and downstream EO performance~\cite{lee2025geosquad}.}
  \label{tab:olm_results_2}
  \centering
  \resizebox{\textwidth}{!}{ 
  \begin{tabular}{l l c c c r r r r r r r}
    \toprule
    \multirow{2}{*}{\textbf{Model}} & \textbf{Geospatial}  & \textbf{Chat} & \textbf{Correct.} & \textbf{Success} & \textbf{Tokens} & \textbf{Total} & \textbf{Agro} & \textbf{Climate} & \textbf{Urban} & \textbf{Forest} & \textbf{Vision} \\
     & \textbf{Agent} &  \textbf{API} & \textbf{Rate}\% & \textbf{Rate}\% & \textbf{Avg (k)} & \textbf{Cost} & $\epsilon$\% & $\epsilon$\% & $\epsilon$ \% & $\epsilon$\% & \textbf{R}\% \\
    \midrule
    \multirow{6}{*}{Llama-3.3-70B} & GeoLLM-Engine (GE) &  & 41.33 & 73.6 & 28.45 & \$7.19 & 15.13 & 13.12 & 12.28 & 11.62 & 82.03 \\
    & GE+Our \texttt{ERR/TRM} & \checkmark & 4.37 & 17.7 & 39.24 & \$2.97 & 60.26 & 54.45 & 47.30 & 44.85 & 63.95 \\
    & Chameleon & \checkmark & 10.54 & 10.0 & -- & -- & -- & -- & -- & -- & -- \\
    & GeoLLM-Squad (GS) &  & 46.02 & 73.3 & 36.60 & \$40.79 & 24.54 & 21.38 & 21.27 & 20.67 & 88.09 \\
    & GS+Our \texttt{ERR/TRM} & \checkmark & 11.95 & 32.3 & 126.60 & \$32.80 & 26.50 & 22.86 & 21.35 & 21.09 & 33.33 \\ \cmidrule(lr){2-12}
    & Geo-OLMs & \checkmark & 31.30 & 72.3 & 81.73 & \$7.57 & 13.70 & 8.92 & 11.17 & 9.94 & 84.00 \\
    \midrule
    \multirow{6}{*}{Llama-3.2-3B} & GeoLLM-Engine (GE) &  & 11.78 & 11.5 & 40.84 & \$0.05 & 32.78 & 29.43 & 27.76 & 27.38 & 48.84 \\
    & GE+Our \texttt{ERR/TRM} & \checkmark & 4.26 & 17.7 & 30.75 & \$0.03 & 22.31 & 19.02 & 17.48 & 16.51 & 17.86 \\
    & Chameleon & \checkmark & 0.70 & 5.9 & -- & -- & -- & -- & -- & -- & -- \\
    & GeoLLM-Squad (GS) &  & 12.65 & 15.4 & 68.80 & \$0.40 & 41.03 & 36.79 & 33.77 & 33.48 & 61.48 \\
    & GS+Our \texttt{ERR/TRM} & \checkmark & 4.42 & 18.7 & 72.80 & \$0.37 & 31.54 & 28.21 & 26.67 & 25.84 & 37.21 \\ \cmidrule(lr){2-12}
    & Geo-OLMs & \checkmark & 3.17 & 18.5 & 75.18 & \$0.40 & 30.26 & 27.18 & 25.64 & 26.41 & 39.47 \\
    \midrule
    \multirow{5}{*}{Mistral-24B} & GeoLLM-Engine (GE) &  & 4.73 & 4.9 & 13.67 & \$0.44 & 75.17 & 70.78 & 66.21 & 65.00 & 86.54 \\
    & GE+Our \texttt{ERR/TRM} & \checkmark & 29.71 & 45.4 & 40.97 & \$0.92 & 25.75 & 22.49 & 21.44 & 22.36 & 61.25 \\
    & GeoLLM-Squad (GS) &  & 21.60 & 5.4 & 98.69 & \$5.44 & 30.00 & 25.79 & 24.21 & 25.00 & 83.62 \\
    & GS+Our \texttt{ERR/TRM} & \checkmark & 14.16 & 8.2 & 68.54 & \$2.63 & 23.85 & 20.62 & 19.33 & 19.85 & 59.81 \\ \cmidrule(lr){2-12}
    & Geo-OLMs & \checkmark & 42.55 & 59.7 & 61.29 & \$1.12 & 11.94 & 9.90 & 9.97 & 10.35 & 29.12 \\
    \midrule
    \multirow{5}{*}{Phi4-14B} & GeoLLM-Engine (GE) &  & 43.73 & 41.5 & 36.49 & \$0.64 & 16.88 & 13.62 & 12.56 & 13.00 & 94.14 \\
    & GE+Our \texttt{ERR/TRM} & \checkmark & 45.25 & 41.5 & 41.09 & \$0.67 & 18.91 & 15.74 & 15.18 & 13.91 & 91.80 \\
    & GeoLLM-Squad (GS) &  & 51.93 & 48.2 & 101.91 & \$5.32 & 21.93 & 18.97 & 18.27 & 18.07 & 89.72 \\
    & GS+Our \texttt{ERR/TRM} & \checkmark & 52.88 & 52.6 & 121.73 & \$5.43 & 18.93 & 15.80 & 14.31 & 14.47 & 85.26 \\ \cmidrule(lr){2-12}
    & Geo-OLMs & \checkmark & 51.30 & 51.8 & 54.72 & \$0.75 & 21.79 & 18.47 & 16.97 & 17.00 & 85.19 \\
    \bottomrule
  \end{tabular}}
\end{table*}

\textbf{Prior work with our self-reflect logic}. To this end, we augment our baseline implementations by applying our chat-based logic (accumulating messages over task progression and chat APIs) along with error correction and termination checks. We refer to these improved baselines as "our \texttt{+ERR/TRM}" in Table~\ref{tab:olms_qwen_results}. These enhancements incorporate all the prompting improvements of our method except for state-driven reasoning. Indeed, we observe that across nearly all models, these refinements lead to higher success rates, bringing performance closer to Geo-OLM. However, GeoLLM-Engine continues to lag, while GeoLLM-Squad remains more expensive due to its complex orchestration logic.

\textbf{Other generalist baselines}. Chameleon remains reasonably competitive in task completion rates, but suffers from high execution costs due to its compositional orchestration overhead and multiple cross-agent communication rounds. Meanwhile, Magentic fails entirely, with completion rates dropping below 5\% for all OLMs, preventing us from collecting meaningful performance metrics. This is consistent with cautionary notes from its developers, who tested it primarily against GPT-4o~\cite{wu2023autogen, fourney2024magentic}. This highlights a broader limitation of applying generalist approaches to geospatial and sustainability studies -- while effective for proprietary LLMs, they often fail to generalize as plug-and-play solutions for more complex and nuanced tasks, particularly with smaller OLMs.

\textbf{OLM vs. GPT trends}. While performance remains competitive at larger scales, the trade-offs become more pronounced as we scale down below 32B. In the 7B–14B range, success rates drop by approximately 20\%, but at a 10$\times$ and 100$\times$ cost reduction compared to GPT-4o-mini and GPT-4o, respectively. As a helpful consideration for practitioners, we note that all models collapse below 3B parameters -- a trend consistent across model families beyond Qwen.

\textbf{Correctness vs. Tokens} Finally, an interesting trend emerges with respect to token consumption: smaller models often use more tokens per task than larger ones. This is a direct result of lower correctness rates, causing agents to repeatedly attempt function calls, leading to excessive retries and inflated token costs. Across all methods and model families, we observe a correlation between lower correctness rates and higher per-task token usage. As motivation for future work, we hypothesize that improving reasoning for better correctness rates (e.g., tool selection) would reduce token overhead, thereby lowering latencies and further reducing costs.

\begin{figure*}[t]
  \centering
  \includegraphics[width=\linewidth]{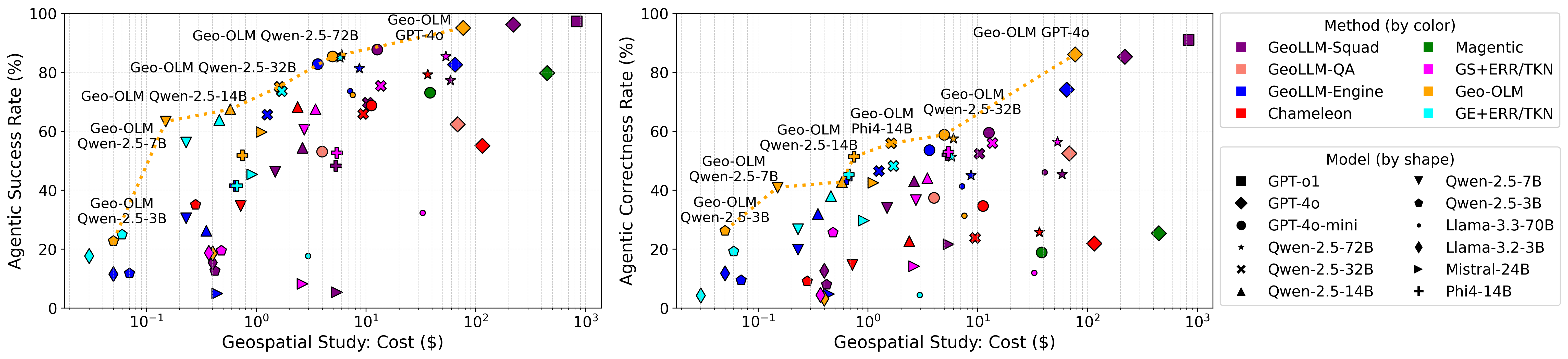}
  \caption{Agentic success and correctness rates vs. benchmarking cost (for a typical benchmark size of 2K user queries) across various models (distinguished by markers) and prompting techniques (distinguished by colors). Overall, Geo-OLM achieves better performance-cost trade-offs across model scales from 72B down to 3B parameters, sustaining agentic performance at reduced computational expenses (best viewed in color).}
  \label{fig:scatters}
  \Description{Agentic performance vs. cost trade-off.}
\end{figure*}

\textbf{Other OLM families - Llama}. We present results for the latest variants of LLaMA, Mistral, and Phi, with most models introduced in recent weeks (Table~\ref{tab:olm_results_2}). For LLaMA 70B, we observe that adding more logic sometimes has the opposite effect, especially with the self-reflection mechanism which surprisingly causes the model to question itself and abandon tasks prematurely. Even more surprisingly, the model performs better without chat API prompting, instead favoring direct text-based parsing, which aligns with reported community findings (e.g., vLLM library~\cite{vllm2025}).

\textbf{Mistral and Phi}. The recently released Mistral appears to handle chat-based and less complex logic more effectively, with both chat API usage and direct text-based parsing working comparably well. However, even with error correction, it struggles with the orchestration burden from GeoLLM-Squad. Meanwhile, Phi demonstrates robust performance but exhibits higher overall token consumption, with noticeably more verbose outputs -- likely due to its reasoning-focused development and pretraining~\cite{abdin2024phi}.

\textbf{OLM and EO tasks}. As expected, we observe a correlation between better agentic performance and improved EO metrics (lower error rates and higher detection recall). Across nearly all model families -- with the exception of smaller 3B variants and Mistral -- Geo-OLMs sustain among the best EO metrics. Finally, we note that detection tasks tend to be more robust to errors, whereas lower success rates have a greater impact on EO product predictions.

\textbf{Discussion}. As captured in Figure~\ref{fig:scatters}, we compare geospatial agent performance across a wide range of proprietary and open language models, spanning scales from 72B down to 3B parameters. Geo-OLMs consistently achieve better trade-offs between performance and cost, maintaining competitive agentic success and correctness rates at significantly lower computational costs. Unlike existing methods, which degrade sharply on smaller OLMs, Geo-OLMs leverage structured state-driven reasoning to mitigating failures such as erroneous tool selection or premature task exits. These results demonstrate that state-driven prompting is an effective strategy for adapting lightweight, community-accessible models to complex geospatial workflows without the prohibitive costs of proprietary LLMs.

\textbf{Cost reduction considerations}. Lower dollar costs directly correspond to reduced runtime per query. That is, within the same model family, a cost reduction by a certain factor corresponds to the runtime being lower by the same factor, reflecting more efficient execution. Across different model families, smaller models benefit from higher parallelization and lower inference latency. We note that it is meaningful to include more hardware-specific criteria~\cite{hu2024routerbench}, such as latency constraints (e.g., data retrieval delays from disk or cloud storage) as well as  energy and power efficiency. As an important direction for future work, we motivate detailed hardware profiling beyond monetary cost, particularly across GPU clusters, edge devices, and local-remote execution setups. A simple extension we are currently investigating is replacing Q4-quantized variants with their full-precision counterparts (as used in Hugging Face baselines~\cite{wolf2019huggingface}). For instance, Q4, Q8, Q16, and full-precision variants are readily available in Ollama~\cite{ollama}, allowing us to flexibly populate the performance-cost trade-off front.

\textbf{Geo-OLMs in Remote sensing}. In Table~\ref{tab:geoengine_results} we present the performance comparison with agentic results and downstream task metrics in remote sensing (RS) satellite-detection workflows~\cite{singh2024geoengine}. Overall, the trends observed in the EO baseline benchmark persist. Geo-OLM maintains strong overall performance. In larger models, detection F1 scores remain particularly high, ranging from 80–90\%, with  success rates remaining high and exceeding 80\%. As expected, downsizing models leads to gradual performance degradation, impacting both agentic reasoning and downstream detection accuracy. Geo-OLM remains competitive across most scales, with the same issues of lower performance persisting for Llama and Mistral models (with success rates dropping to less than 40\%).

\begin{table*}
  \caption{Agentic performance in RS tasks~\cite{singh2024geoengine} across different model families, evaluating correctness, success rates, and downstream task accuracy (Land-coverage classification accuracy and detection F1 scores).}
  \label{tab:geoengine_results}
  \centering
  \resizebox{0.8\textwidth}{!}{ 
  \begin{tabular}{l l c c c c}
    \toprule
    \textbf{Model} & \textbf{Geospatial Agent }& \textbf{Correctness Rate} (\%) & \textbf{Success Rate} (\%) & \textbf{LCC Acc} (\%) & \textbf{Det F1} (\%) \\
    \midrule
    \multirow{3}{*}{GPT-4o} & GeoLLM-Engine & 59.06 & 81.97 & 95.45 & 75.37 \\
     & GeoLLM-Squad & 86.04 & 84.70 & 96.28 & 86.90 \\ \cmidrule(lr){2-6}
     & Geo-OLM & 92.14 & 95.63 & 95.54 & 97.44 \\
    \midrule
    \multirow{3}{*}{GPT-4o-mini} & GeoLLM-Engine & 59.98 & 84.70 & 74.49 & 62.10 \\
     & GeoLLM-Squad & 42.89 & 87.29 & 90.54 & 69.43 \\ \cmidrule(lr){2-6}
     & Geo-OLM & 64.54 & 86.89 & 85.30 & 76.06 \\
    \midrule
    \multirow{3}{*}{Qwen-2.5-72B} & GeoLLM-Engine & 45.75 & 85.12 & 94.72 & 95.96 \\
     & GeoLLM-Squad & 42.60 & 87.50 & 95.26 & 96.45 \\ \cmidrule(lr){2-6}
     & Geo-OLM & 54.59 & 84.70 & 91.25 & 92.28 \\
    \midrule
    \multirow{3}{*}{Qwen-2.5-32B} & GeoLLM-Engine & 54.13 & 62.84 & 95.14 & 90.67 \\
     & GeoLLM-Squad & 53.65 & 65.57 & 91.81 & 88.01 \\ \cmidrule(lr){2-6}
     & Geo-OLM & 55.64 & 73.22 & 94.80 & 85.80 \\
    \midrule
    \multirow{3}{*}{Qwen-2.5-14B} & GeoLLM-Engine & 35.84 & 33.33 & 96.36 & 81.23 \\
     & GeoLLM-Squad & 46.77 & 68.35 & 94.01 & 91.38 \\
     & GeoLLM-Squad & 49.53 & 74.42 & 90.01 & 89.11 \\  \cmidrule(lr){2-6}
     & Geo-OLM & 41.94 & 65.75 & 92.61 & 88.89 \\
    \midrule
    \multirow{3}{*}{Qwen-2.5-7B} & GeoLLM-Engine & 16.51 & 50.38 & 79.70 & 69.93 \\
    & GeoLLM-Squad & 34.09 & 58.02 & 67.51 & 51.11 \\\cmidrule(lr){2-6}
    & Geo-OLM & 37.25 & 59.89 & 67.79 & 71.91 \\
    \midrule
    \multirow{3}{*}{Llama-3.3-70B} & GeoLLM-Engine & 7.25 & 20.45 & 70.83 & 63.75 \\
     & GeoLLM-Squad & 40.59 & 68.48 & 97.20 & 88.45 \\ \cmidrule(lr){2-6}
     & Geo-OLM & 27.09 & 61.54 & 48.78 & 81.28 \\
    \midrule
    \multirow{3}{*}{Llama-3.2-3B} & GeoLLM-Engine & 9.32 & 21.52 & 94.47 & 89.17 \\
     & GeoLLM-Squad & 5.65 & 28.57 & 86.37 & 93.88 \\ \cmidrule(lr){2-6}
     & Geo-OLM & 22.40 & 25.93 & 95.05 & 80.95 \\
    \midrule
    \multirow{3}{*}{Mistral-24B} & GeoLLM-Engine & 8.56 & 15.24 & 83.16 & 85.87 \\
     & GeoLLM-Squad & 30.02 & 10.56 & 93.71 & 84.73 \\ \cmidrule(lr){2-6}
     & Geo-OLM & 34.26 & 36.61 & 59.95 & 88.85 \\
    \midrule
    \multirow{3}{*}{Phi4-14B} & GeoLLM-Engine & 48.05 & 52.20 & 93.45 & 94.72 \\
     & GeoLLM-Squad & 49.53 & 74.42 & 90.01 & 89.11 \\  \cmidrule(lr){2-6}
     & Geo-OLM & 48.96 & 72.41 & 93.22 & 85.20 \\
    \bottomrule
  \end{tabular}
  }
\end{table*}

\section{Case Study: Earthquake Impact}

A possible limitation of state-driven prompting is its reliance on explicitly defined workflow states, raising an important question: does improving OLM performance come at the cost of generalizability? To evaluate this, we conduct a case study simulating how a geospatial analyst could apply our method to a real-world disaster impact assessment. We focus on the February 2023 Turkey earthquake, leveraging a previously released post-disaster satellite and population analysis study~\cite{robinson2023turkey}. This study is an ideal test case for agentic reasoning, as it explicitly describes an analytical workflow with a step-by-step methodology. Crucially, it provides structured datasets (satellite-detected building damage and population density estimates) and tools~\cite{microsoft_globalml}, enabling programmatic verification of agentic performance with high fidelity by matching its findings.

\textbf{Methodology}. To enable agentic evaluation, we convert the study’s datasets and documentation into a vector database, allowing for retrieval-augmented prompting of relevant information. Using GPT, we generate 120 representative question variants that a geospatial analyst might ask based on the study’s findings. Consider a sample question: ``How many buildings were damaged in Nurd\u{a}\u{g}{\i} on February 9?'' or ``What was the estimated impacted population in \.{I}slahiye on February 7?'' These can be programmatically answered from the dataset, providing a direct ground truth for evaluating agentic correctness. This allows us to establish a gold-standard benchmark, by deriving expected answers and matching directly with the study's conclusions. 

Next, we generate agentic ground truths, formatted as sequences of tool-calling steps.We follow the Tool-QA methodology~\cite{zhuang2023toolqa} and use GPT-4o as an oracular solver, providing both the sample question and the programmatically derived solution. The oracle executes API calls to retrieve results, recording step-by-step function calls and argument selections. We discard agentic solutions that do not match the programmatic solution, allowing for a 10\% variance to account for rounding differences in tool calls (typically coordinates).

Finally, we formulate the task as a state-driven workflow tailored to this study, defining states as: \texttt{Init} $\rightarrow$ \texttt{Load} $\rightarrow$ \texttt{Filter} $\rightarrow$ \texttt{Detect} $\rightarrow$ \texttt{Correlate (Building/Population)} $\rightarrow$ \texttt{Plot/Answer} $\rightarrow$ \texttt{End}. The Correlation step reflects the study's implementation, where population estimates are spatially correlated with detected building damage to quantify affected populations. We expand Geo-OLM with the tools for satellite-based damage detection and urban population estimation~\cite{microsoft_globalml}. We convert the study data to DataFrames for frontend integration with our framework. For evaluation, we adopt the same success rate criteria as in the baseline benchmarks. 

\begin{table}
  \caption{Agentic performance across model families towards replicating the February 2023 Turkey earthquake impact study~\cite{robinson2023turkey}, evaluating correctness, success rates, and cost.}  
  \label{tab:earthquake_results}
  \centering
  \resizebox{\columnwidth}{!}{ 
  \begin{tabular}{l l c c c}
    \toprule
    \multirow{2}{*}{\textbf{Model}} & \textbf{Geospatial} & \textbf{Correct.} & \textbf{Success}  & \textbf{Total}  \\
     & \textbf{Agent} & \textbf{Rate}\% & \textbf{Rate}\% & \textbf{Cost} (120)  \\
    \midrule
    \multirow{4}{*}{GPT-4o} & GE+\texttt{ERR/TRM} & 62.42\% & 100.0\% & \$2.14 \\
    & Chameleon & 88.52\% & 100.0\% & \$12.14 \\
    & GS+\texttt{ERR/TRM} & 95.92\% & 100.0\% & \$9.99 \\ \cmidrule(lr){2-5}
    & Geo-OLM & 90.52\% & 100.0\% & \$2.59 \\ 
    \midrule
    \multirow{3}{*}{GPT-4o-mini} & GE+\texttt{ERR/TRM} & 58.43\% & 100.0\% & \$0.11 \\
    & GS+\texttt{ERR/TRM} & 89.44\% & 100.0\% & \$0.59 \\ \cmidrule(lr){2-5}
    & Geo-OLM & 80.72\% & 100.0\% & \$0.28 \\
    \midrule
    \multirow{4}{*}{Qwen-2.5-72B} & GE+\texttt{ERR/TRM} & 58.12\% & 100.0\% & \$0.24 \\
    & Chameleon & 68.66\% & 98.3\% & \$3.16 \\
    & GS+\texttt{ERR/TRM} & 80.94\% & 100.0\% & \$2.31 \\ \cmidrule(lr){2-5}
    & Geo-OLM & 78.67\% & 100.0\% & \$0.29 \\
    \midrule
    \multirow{4}{*}{Qwen-2.5-32B} & GE+\texttt{ERR/TRM} & 43.65\% & 96.7\% & \$0.06 \\
    & Chameleon & 44.56\% & 66.7\% & \$0.54 \\
    & GS+\texttt{ERR/TRM} & 62.04\% & 96.7\% & \$0.26 \\ \cmidrule(lr){2-5}
    & Geo-OLM & 62.59\% & 98.3\% & \$0.08 \\
    \midrule
    \multirow{3}{*}{Qwen-2.5-14B} & GE+\texttt{ERR/TRM} & 40.11\% & 45.0\% & \$0.02 \\
    & GS+\texttt{ERR/TRM} & 49.22\% & 98.3\% & \$0.15 \\ \cmidrule(lr){2-5}
    & Geo-OLM & 43.16\% & 90.4\% & \$0.06 \\
    \midrule
    \multirow{3}{*}{Qwen-2.5-7B} & GE+\texttt{ERR/TRM} & 35.67\% & 30.0\% & \$0.01 \\
    & GS+\texttt{ERR/TRM} & 46.00\% & 36.7\% & \$0.08 \\ \cmidrule(lr){2-5}
    & Geo-OLM & 50.40\% & 38.3\% & \$0.02 \\
    \midrule
    \multirow{3}{*}{Llama-3.3-70B} & GeoLLM-Engine (GE) & 38.23\% & 95.0\% & \$0.20 \\
    & GeoLLM-Squad (GS) & 75.35\% & 100.0\% & \$1.49 \\ \cmidrule(lr){2-5}
    & Geo-OLM & 70.58\% & 100.0\% & \$0.34 \\
    \midrule
    \multirow{2}{*}{Phi4-14B} & GS+\texttt{ERR/TRM} & 66.38\% & 50.0\% & \$0.19 \\
    & Geo-OLM & 41.84\% & 65.0\% & \$0.02 \\
    \bottomrule
  \end{tabular}
  }
\end{table}

\textbf{Results}. We observe that trends persist in this case study (Table~\ref{tab:earthquake_results}), with Geo-OLM maintaining good performance-cost trade-off across nearly all model versions. As model size decreases, correctness rates drop even for this simpler task, likely due to increased typos and argument formatting errors. However, success rates remain consistently high, reaching 100\% in larger models, as these tasks follow a simpler single-scope workflow, unlike the more complex multi-agent, multi-scope flows in the baseline benchmark.

\section{Limitations and Future Work}

One key limitation of state-driven reasoning is its reliance on domain expertise to construct structured workflows. Designing effective prompts requires a clear understanding of the task and careful engineering of workflow states~\cite{wu2024stateflow}, which may introduce a barrier to adoption. However, flow-based tooling is already a common practice in LLM deployment, particularly within UI-driven frameworks such as AutoGen Studio~\cite{dibia2024autogen, wu2023autogen} and Azure Prompt Flow~\cite{azure2024promptflow}. From this perspective, state-driven reasoning can be seen as an extension of existing deployment paradigms rather than an entirely new requirement. This also presents an opportunity for future work to explore human-computer interaction (HCI) practices for integrating structured reasoning into interactive agent development and enhancing interfaces beyond tool invocation.

Another consideration is the rapid improvement of smaller language models, which may eventually narrow the performance-cost gap. While this trend is undeniable, our results show that even commercial solutions do not yet achieve human-like performance in complex geospatial workflows. Since both proprietary and smaller open models are primarily optimized for general-purpose reasoning, there will continue to be a need for domain-specialized agents in fields such as geosciences~\cite{szwarcman2024prithvi} and biosciences~\cite{lobentanzer2025biomedical}. In this context, state-driven prompting remains valuable as a method to enhance execution reliability and reasoning efficiency, even as cost differences diminish over time.

\section{Conclusion}

We introduced Geo-OLM, a geospatial agentic framework based on state-driven prompting, which explicitly separates reasoning from tool execution. Through comprehensive evaluations on geospatial benchmarks and a domain-specific case study, we demonstrated that Geo-OLMs achieve competitive performance, staying within 10–20\% of proprietary models while reducing costs by 10–100 $\times$. These results highlight the viability of structured agentic workflows for cost-efficient, scalable geospatial AI.

\bibliographystyle{ACM-Reference-Format}
\bibliography{geoslm}

\end{document}